\documentclass[letter, 10 pt, conference]{ieeeconf}

\IEEEoverridecommandlockouts                              

\usepackage{xcolor}
\usepackage{graphicx}
\usepackage{xspace}
\usepackage{listings}
\usepackage{amsmath}
\usepackage{amsfonts}
\usepackage{todonotes}
\usepackage{tabularx}
\usepackage{caption}
\usepackage{subcaption}
\usepackage{algorithm}
\usepackage{algorithmic}
\usepackage{cite}
\usepackage{hyperref}
\usepackage[capitalize]{cleveref}

\usepackage{mathrsfs}
\usepackage{bbm}

\title{\LARGE \bf
Adaptive Failure Search Using Critical States\\ from Domain Experts
}

\author{Peter Du and Katherine Driggs-Campbell
\thanks{P. Du and K. Driggs-Campbell are with the Department of Electrical and Computer Engineering at the University of Illinois at Urbana-Champaign. email: \{peterdu2,krdc\}@illinois.edu}%
}

\begin{document}

\maketitle


\begin{abstract}

Uncovering potential failure cases is a crucial step in the validation of safety critical systems such as autonomous vehicles. Failure search may be done through logging substantial vehicle miles in either simulation or real world testing. Due to the sparsity of failure events, naive random search approaches require significant amounts of vehicle operation hours to find potential system weaknesses. As a result, adaptive searching techniques have been proposed to efficiently explore and uncover failure trajectories of an autonomous policy in simulation. Adaptive Stress Testing (AST) is one such method that poses the problem of failure search as a Markov decision process and uses reinforcement learning techniques to find high probability failures. However, this formulation requires a probability model for the actions of all agents in the environment. In systems where the environment actions are discrete and dependencies among agents exist, it may be infeasible to fully characterize the distribution or find a suitable proxy. This work proposes the use of a data driven approach to learn a suitable classifier that tries to model how humans identify \textit{critical states} and use this to guide failure search in AST. We show that the incorporation of critical states into the AST framework generates failure scenarios with increased safety violations in an autonomous driving policy with a discrete action space. 

\end{abstract}


\section{Introduction}


Autonomous systems are rapidly becoming tangible technologies that are deployed into the real world. In safety critical applications like autonomous driving, extensive testing must be performed prior to deployment in order to ensure the safety of both the humans using the technology and the vulnerable participants in the environment. Validation of autonomous vehicles (AVs) continues to pose challenges for the designers of these systems due to the diverse range of scenarios vehicles may face when operating on public roads. Physical vehicle testing is a commonly employed method to gain confidence in an AV's ability to function safely in the real world \cite{huang2018testing}; however, the monetary and time costs associated with real world vehicle testing make it difficult to extensively validate all scenarios an AV may face \cite{Kalra2014}. Simulation may be used to complement physical vehicle testing by allowing AV designers to run a large fleet of vehicles for long periods of time \cite{Koopman2016challenges}. 

However, even with the increased testing capacity afforded by simulation, naive random search techniques may fail to yield adequate coverage of the failure space due to the rarity of dangerous scenarios. As a result, various approaches have been explored to generate critical scenarios that may lead to AV failure. Deep generative models such as \cite{ding2020multimodal} and \cite{ding2019new} make use of a flow-based and VAE model, respectively, to produce informative driving scenarios. Adaptive sampling approaches have been proposed to search for critical scenarios that are likely to result in AV failure \cite{MULLINS2018197,OKelly2018ScalableEA,8500421,7571159}.

Adaptive Stress Testing (AST) is a technique that uses simulation to find failures in autonomous systems \cite{lee2015adaptive}. The search for vulnerabilities is posed as finding the most probable failures of the system and is formulated as a Markov decision process (MDP). RL techniques are used to adaptively sample the environment and solve the MDP. In the standard formulation, the reward function of AST depends on the log-likelihood of the state transitions within the simulator and assumes environment actions are IID, thus optimizing a policy to generate \textit{most likely} failures. The RL solver samples actions from a continuous space and the Mahalanobis distance is often used as a proxy for the probability of actions \cite{mahalanobis}. However, this formulation is not well suited for the validation of high-level decision making systems with discrete action spaces where a similar proxy is not readily available or may need to be hand crafted. In these instances, we may wish to seek a policy that generates worse case or risk sensitive failures. To do so, a training heuristic can be used in place of the likelihood in the reward function \cite{koren2019astformulation}, but this poses difficulties in the design of a suitable heuristic without limiting the scope of failures. 

We seek to address this problem by foregoing a manually crafted heuristic and instead consider the characterization of \textit{critical states}. We postulate that failure scenarios may be attributed to the combined affects of \textit{unsafe} states and that scenarios where the agent exhibits an increased amount of this unsafe behaviour is informative for safety validation. Several metrics exist for characterizing the criticality of a robot's state such as looking at its risk measure or distribution of utility values \cite{majumdar2017robot,huang2018}. We treat the identification of critical states as a classification task and employ a data driven approach to learn a suitable classifier that predicts whether the state of the system is dangerous or not. The use of this learned model allows AST to efficiently generate failure trajectories with increased unsafe behaviour for black box systems with a discrete high-level action space. 

We present the following contributions:
\begin{enumerate}
    \item We discuss the limitations of AST when applied to systems with discrete action spaces and propose a framework incorporating critical states to better identify meaningful failures. 
    \item We consider two characterizations of critical states. The first defines criticality based solely on the observations and policy of the autonomous agent. The second considers the perspective of an outside onlooker and uses observations from the environment as a whole. 
    \item We apply our methods of failure generation on an autonomous vehicle policy that is trained to output discrete driving decisions. We show the effectiveness of the approach in capturing scenarios with a greater amount of AV misbehaviour and dangerous actions. 
\end{enumerate}

This paper is organized as follows: \cref{sec:background} gives an overview of AST and the characterization of critical states in an RL policy. \cref{sec:methods} describes the incorporation of a learning-based critical state classifier into the AST framework. \cref{sec:experiments,,sec:results} present the experiments and results of validating a DQN autonomous driving policy. We conclude and discuss future directions in \cref{sec:conclusion}. 

\section{Background}
\label{sec:background}


This section provides an overview on the AST methodology and characterization of critical states. 

\subsection{Adaptive Stress Testing}
\label{sec:backgroundAST}
Adaptive Stress Testing (AST) is a technique that poses the problem of validation and failure search as an MDP and then applies reinforcement learning methods to generate a policy that finds likely failure modes of the system-under-test (SUT). The modularity of AST allows one to use various different solvers and apply the technique to validate a range of systems. In prior studies, AST has been applied to both aircraft collision avoidance and autonomous driving policies \cite{lee2015adaptive, koren2018adaptive}.

\begin{figure}[!h]
    \centering
    \hspace*{0.4cm}
    \includegraphics[width=0.95\columnwidth]{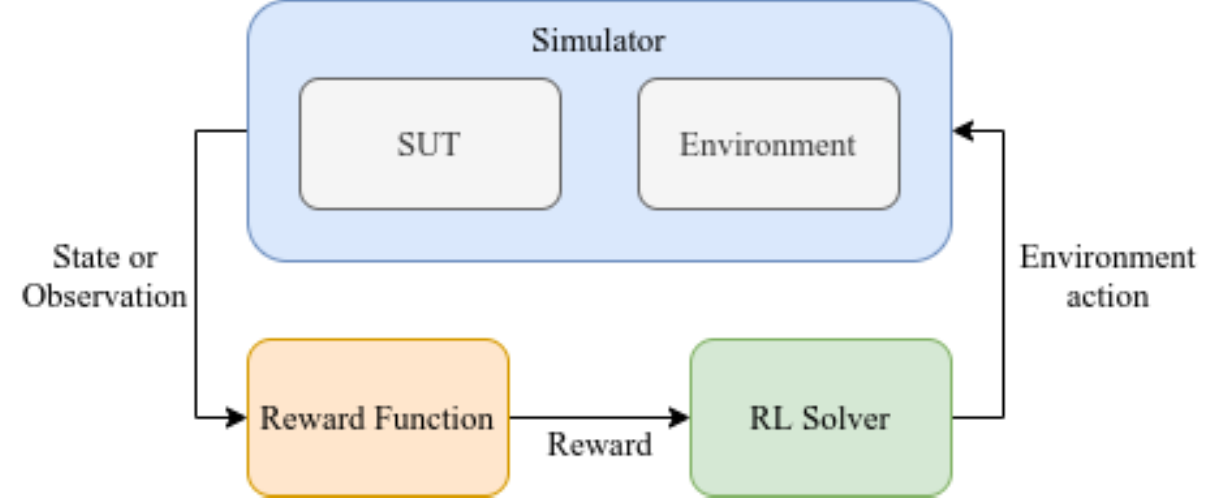}\\
    \caption{The Adaptive Stress Testing architecture.}
    \label{fig:ast_block}
\end{figure}

AST is made up of three main component blocks consisting of a simulator $\mathscr{S}$, reward function $R$, and RL solver $S$. A diagram of the architecture is shown in \cref{fig:ast_block}. The simulator $\mathscr{S}$ contains both the SUT and the environment that it acts in. A set of environment actions $\mathcal{A}$ is used to step the simulator forward in time. In the case of an autonomous vehicle application, the SUT could be the ego-vehicle under autonomous control and the environment actions may consist of sensor noise on the AV, acceleration/braking of surrounding vehicles, motion of pedestrians, etc. A set of goal states $E$ is defined for the simulator and represents failures of the SUT. In our AV scenario, this contains all states where the ego-vehicle collides with another agent.

The RL solver $S$ generates environment actions which are used to update the state of the simulator $\mathscr{S}$. AST allows $\mathscr{S}$ to be black-box or semi-black-box and does not require the simulator to expose its complete internal state as long as it satisfies the following interface:

\begin{itemize}
    \item \verb|Init|$(\mathscr{S})$: Reset the simulator to its starting state.
    \item \verb|Step|$(\mathscr{S}, a)$: Execute action $a$ and step the simulator forward in time. Return an indicator of whether the new state of the simulator is in the set of goal states $E$.
    \item \verb|IsTerminal|$(\mathscr{S})$: Check if the current state is in $E$ or if the time horizon has been reached.
\end{itemize}

Once in its new state, the simulator passes any relevant state/observation information to the AST reward function $R$ which generates a reward signal. The reward is returned to the solver and the policy is updated before generating the next environment action. The reward function is defined as:

\begin{equation}
\nonumber
\label{eq:ast_reward}
R\left(s\right) = \left\{
        \begin{array}{ll}
            0 &  s \in E \\[7pt]
            -\alpha - \beta f(s) &  s \notin E, t\geq T \\[7pt]
            -g(a) - \eta h(s) &  s \notin E, t < T
        \end{array}
    \right.
\end{equation}

\noindent where:
\begin{itemize}
    \item $s$ is the current state/observation of the simulator.
    \item $E$ is the set of goal states (system failures).
    \item $T$ is the time horizon of simulation.
    \item $\alpha$ is a constant to penalize trajectories that do not end in failure.
    \item $\beta f(s)$ is an optional heuristic to distinguish among trajectories that do not end in failure.
    \item $g(a)$ is the reward for the current action (proportional to the log probability).
    \item $\eta h(s)$ is a heuristic reward for each training step. 
\end{itemize}

The AST framework allows for the use of various RL solvers. In this work, we use Monte Carlo Tree Search (MCTS) as our RL policy \cite{Browne2012}. We have seen success in prior works with MCTS in finding failure trajectories when used with AST \cite{lee2018differential, koren2018adaptive, 8917242}. 

\vspace{0.4cm}

\subsection{Critical State Classification}
\label{sec:backgroundCS}
Learning based approaches have been used to generate complex policies for autonomous systems. For example, deep reinforcement learning (DRL) may be used to perform an autonomous driving task or a neural network may be trained to perform a visuomotor task. \cite{kiran2020deep, levine2016endtoend}. Often in these black box systems, the neural networks are trained end-to-end making it difficult to fully characterize the system's behaviour as it is infeasible to examine the precise actions of the autonomous agent in all possible states. To address this, Huang et al. propose that there exists a smaller set of \textit{critical states} which we can examine\cite{huang2018}. Here, critical states are ones where the choice of agent actions has a significantly larger affect on the expected utility value of the outcome. Defined more concretely, critical states can be characterized by states where the maximum action-value pair (Q-value) is greater than the average action-value pair by some threshold $t$. Intuitively, this definition views critical states as scenarios in which the agent's policy believes that acting randomly will perform much worse than acting optimally. 

If the SUT consists of a policy that learns a value or action-value pair, we may use this form of \textit{Q-value based critical states} in the AST reward function in lieu of a hand crafted heuristic. However, we suggest that such a definition of critical states may not be ideal for failure trajectory search as not all results may be dangerous. For example, in an autonomous driving scenario where the ego vehicle is following a car with additional cars to the left and right, the optimal action (staying at a constant speed) may be the significantly higher action-value pair as other actions lead to a collision (lane change left, right, accelerate, etc). Using the value based characterization of critical states, such a scenario would be labeled as critical. However, given that the following distance was acceptable, such a state may not be \textit{dangerous}. Likewise, in a scenario where collision is imminent, all actions in the state may have a similar action-value pair as the choice of action has little effect on the outcome of the scenario. In such situations, the state would not be classified as critical even though the scenario is dangerous and should be appropriately weighted in the AST reward function.

To capture the appropriate critical states useful for failure scenario identification, we suggest that criticality should be defined not just from the SUT's perspective, but also incorporate features from the rest of the environment. To accomplish this, we propose learning directly from human observations to identify critical states. 

\section{Methods}
\label{sec:methods}


This section describes the task of identifying critical states and its integration into the AST reward function. 

\subsection{Human based Critical State (HCS) Classifier} 

Training data for the HCS classifier can be collected from trajectories obtained through random simulation or sample trajectories from AST with a purely heuristic based reward function. Usage of AST with heuristic reward can help improve data collection efficiency and generate an increased number of failure/dangerous states. After data collection, each trajectory is partitioned into a set of states and labelled via human response. States may be represented by vectors containing the simulators internal state (in the case of a white-box/semi-black-box simulator), the ego vehicle's observation, or a mixture of both. The labelled data is split into sets containing only positive and negative samples. The two sets are randomly shuffled and merged to include a balanced amount of positive and negative labels. 

The training dataset $D = \{(s_1,l_1), (s_2,l_2),... (s_n,l_n)\}$ consists of state/label pairs where:

\begin{itemize}
    \item $s_i \in \mathbb{R}^d$ is a state that contains both the environment actions (as defined in \cref{sec:backgroundAST}), and the ego vehicle's observations/actions.
    \item $l_i \in \{0,1\}$ is a human labelled indicator of whether the scenario is dangerous.
\end{itemize}
Given a dataset $D$, we learn a mapping $\mathcal{F}: \mathbb{R}^d \rightarrow [0,1]$ which predicts a score on each state. We treat this as a soft classification task and fit a Bayesian neural network over the training dataset using the binary cross entropy loss. The network takes in a feature vector of the state $s$ and outputs a 2x1 vector where the entries correspond to the probability of the state being either dangerous or not. The use of soft classification and a Bayesian network allows us to characterize the uncertainty of predictions given by the HCS classifier and incorporate it into the AST reward function.

\begin{figure}[!t]
    \centering
    \hspace*{-0.1cm}
    \includegraphics[width=\columnwidth]{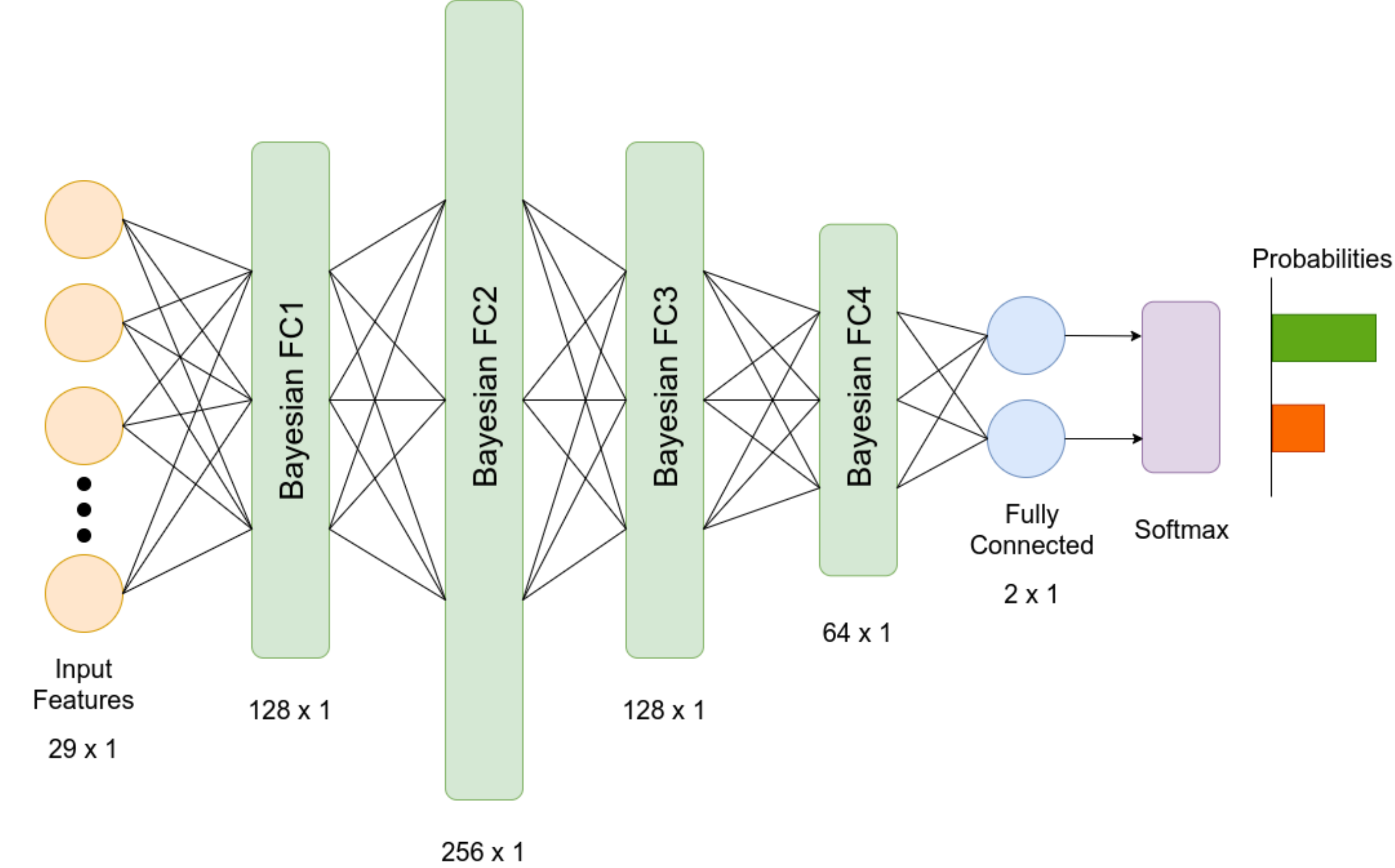}\\
    \caption{Network architecture of HCS classifier.}
    \label{fig:hcs_arch}
\end{figure}




\subsection{AST Reward Function}
\label{sec:dsc_reward}
Once trained, we can use the HCS classifier to infer a probability distribution of a given state being predicted as dangerous. During failure search, each feature vector corresponding to the current state $s$ is passed through the Bayesian network $n$ times. Let $\mu$ and $\sigma^2$ be the mean and variance of the distribution obtained. The AST reward function evaluated for a state $s$ at time $t$ is then given as follows:

\begin{equation}
\nonumber
\label{eq:ast_reward}
R_{hcs}\left(s\right) = \left\{
        \begin{array}{ll}
            0 &  s \in E\\[7pt]
            -\infty &  s \notin E, t\geq T \\[7pt]
            (1-\sigma^2)\beta\mu+ \sigma^2h(s) &  s \notin E, t < T
        \end{array}
    \right.
\end{equation}
\vspace{0.3cm}

Here, $E$ is the set of user defined failure states and $T$ is a finite time horizon. $h(s)$ is a heuristic measure given by the longitudinal distance between the ego vehicle and the closest vehicle in the same lane ahead of it. $\beta$ is a user adjustable constant to tune the weight of the reward given by critical states and should be matched to be of a similar magnitude to the heuristic measure. 

The reward uses the variance of the predicted distribution as a proxy for the confidence of the HCS classifier in order to adjust the contribution of the reward obtained from the heuristic versus the classifier. This hybrid form of reward function allows us to compensate for scenarios where the classifier has low confidence in the validity of its prediction. In these cases, the weight of the HCS classifier is diminished and does not feed the RL solver a misguided reward signal. 

\section{Experiments}
\label{sec:experiments}


This section describes the experimental setup and scenario used to evaluate the performance of AST with HCS reward and various baselines. 

\begin{figure}[t]
    \centering
    \includegraphics[width=.85\columnwidth]{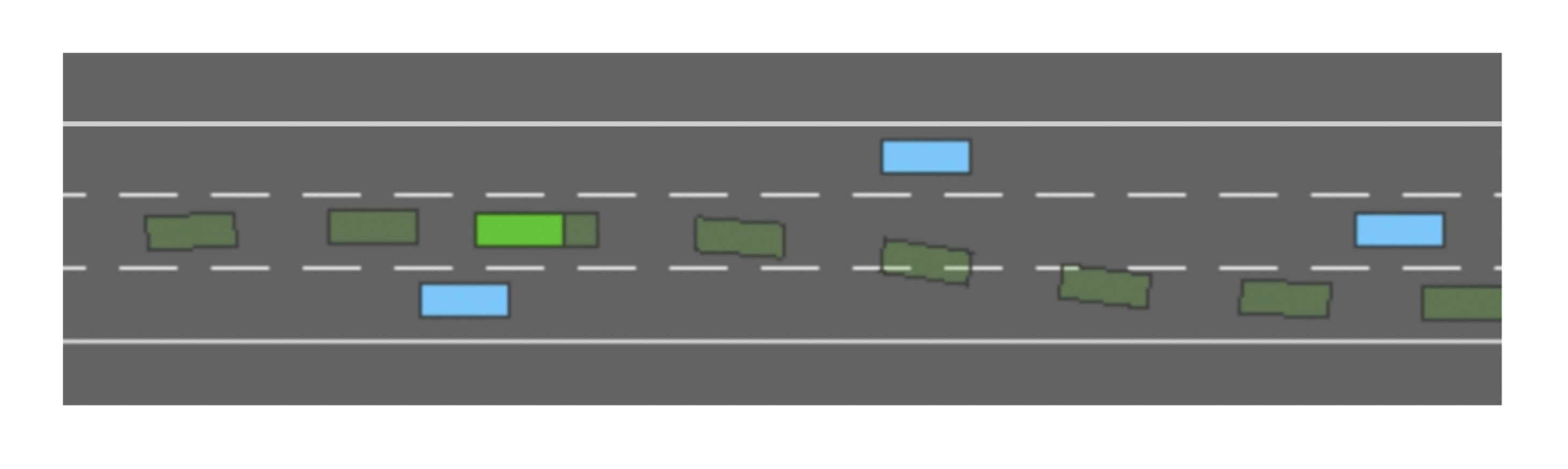}\\
    \caption{Autonomous vehicle highway simulator \cite{highway-env}.}
    \label{fig:highway_env}
\end{figure}

The scenario used to test the AV consists of a three lane straight highway with other vehicles on the road driving at various speeds. The Highway-Env OpenAI Gym environment is used to train a vehicle controller whose goal is to achieve a high speed while avoiding collision with surrounding vehicles \cite{highway-env}. A Deep Q Network (DQN) vehicle controller is trained using the Stable-Baselines library \cite{stable-baselines}. An example of the simulator environment is shown in \cref{fig:highway_env}.

An AST interface is built around the Highway-Env Gym environment and exposes the required components as discussed in \cref{sec:backgroundAST}. Here, the SUT is the autonomous vehicle that is guided by the DQN controller. The action space of the vehicles (both ego and environment) consists of five high level actions: left lane change, right lane change, idle, accelerate, and brake. During failure search, AST has direct control over the environment actions which dictate the behaviour of the eight closest vehicles surrounding the ego AV. Each environment action is represented by an 8-tuple: 
$$
a_{env} = (a^{(1)}, a^{(2)}, a^{(3)},... a^{(8)})
$$
\noindent where $a^{(i)}$ is the high level action for the $i$th closest environment vehicle. 

The state that is used as input to the HCS classifier is taken to be a 3-tuple:
$$
s = (a_{ego}, a_{env}, o_{ego})
$$

\noindent where $a_{ego}$ is the action of the ego vehicle, $a_{env}$ is the environment action, and $o_{ego}$ is the observation of the ego vehicle. This observation is the same one used by the DQN controller and consists of the $x$ position, $y$ position, $x$ velocity, and $y$ velocity of the five closest vehicles surrounding the ego AV. 

We run AST using a simulator initialized with 40 vehicles and an MCTS solver to find failure scenarios of the DQN driving policy. Two baselines are considered. The first uses a reward function that relies on a heuristic measure $h(s)$ (as defined in \cref{sec:dsc_reward}) to train the solver. The reward is given as:

\begin{equation}
\nonumber
\label{eq:heuristic_reward}
R_{heur}\left(s\right) = \left\{
        \begin{array}{ll}
            0 &  s \in E\\[7pt]
            -\infty &  s \notin E, t\geq T \\[7pt]
            h(s) &  s \notin E, t < T
        \end{array}
    \right.
\end{equation}
\vspace{0.3cm}

The second uses a Q-value critical state criteria as described in \cref{sec:backgroundCS}. The reward function is given as:

\begin{equation}
\nonumber
\label{eq:qcs_reward}
R_{qcs}\left(s\right) = \left\{
        \begin{array}{ll}
            0 &  s \in E\\[7pt]
            -\infty &  s \notin E, t\geq T \\[7pt]
            \max\limits_{a}Q(s,a) - \frac{1}{|\mathcal{A}|}\sum\limits_a Q(s,a) &  s \notin E, t < T
        \end{array}
    \right.\\
\end{equation}
\vspace{0.3cm}

We compare these baselines with the HCS reward defined in \cref{sec:dsc_reward} that uses a trained classifier. 


\begin{figure}[b]
    \centering
    \includegraphics[width=\columnwidth]{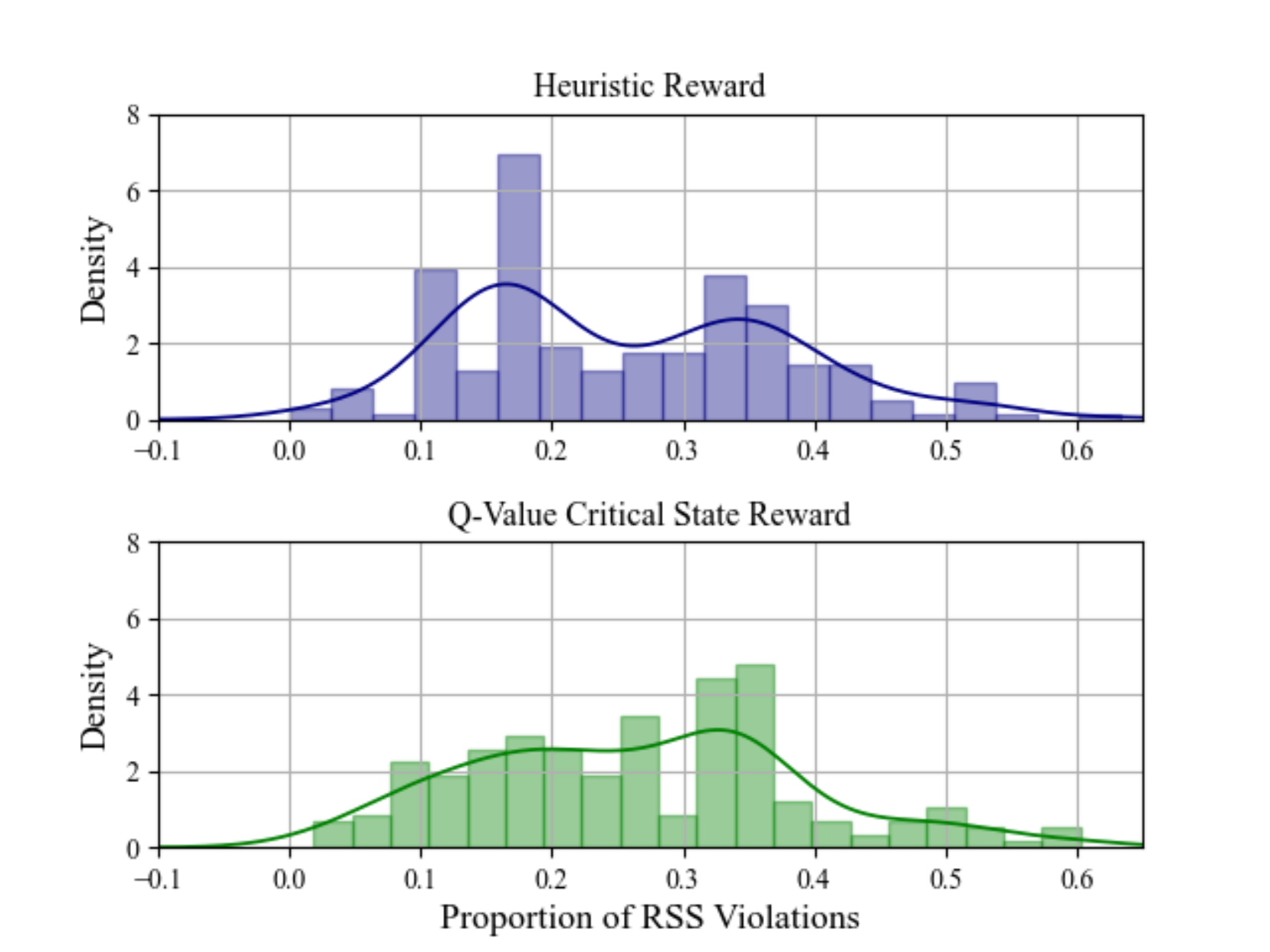}\\
    \caption{Normalized histograms of proportion of RSS violations for failure trajectories found using a Q-value critical state metric.}
    \label{fig:RSS_hist_CS} 
\end{figure}

\section{Results}
\label{sec:results}


In this section we demonstrate the results of using AST to validate the autonomous driving agent shown in \cref{sec:experiments}. We evaluate the quality of the results using Responsibility-Sensitive Safety (RSS): a set of interpretable mathematical models for driving safety assurance \cite{shalev2017formal}. For each reward, we perform an analysis of the failure trajectories found to identify the proportion of states deemed unsafe by RSS. We also consider a qualitative assessment of the results.



\begin{figure}[t]
    \centering
    \hspace*{-0.5cm}
    \includegraphics[width=0.9\columnwidth]{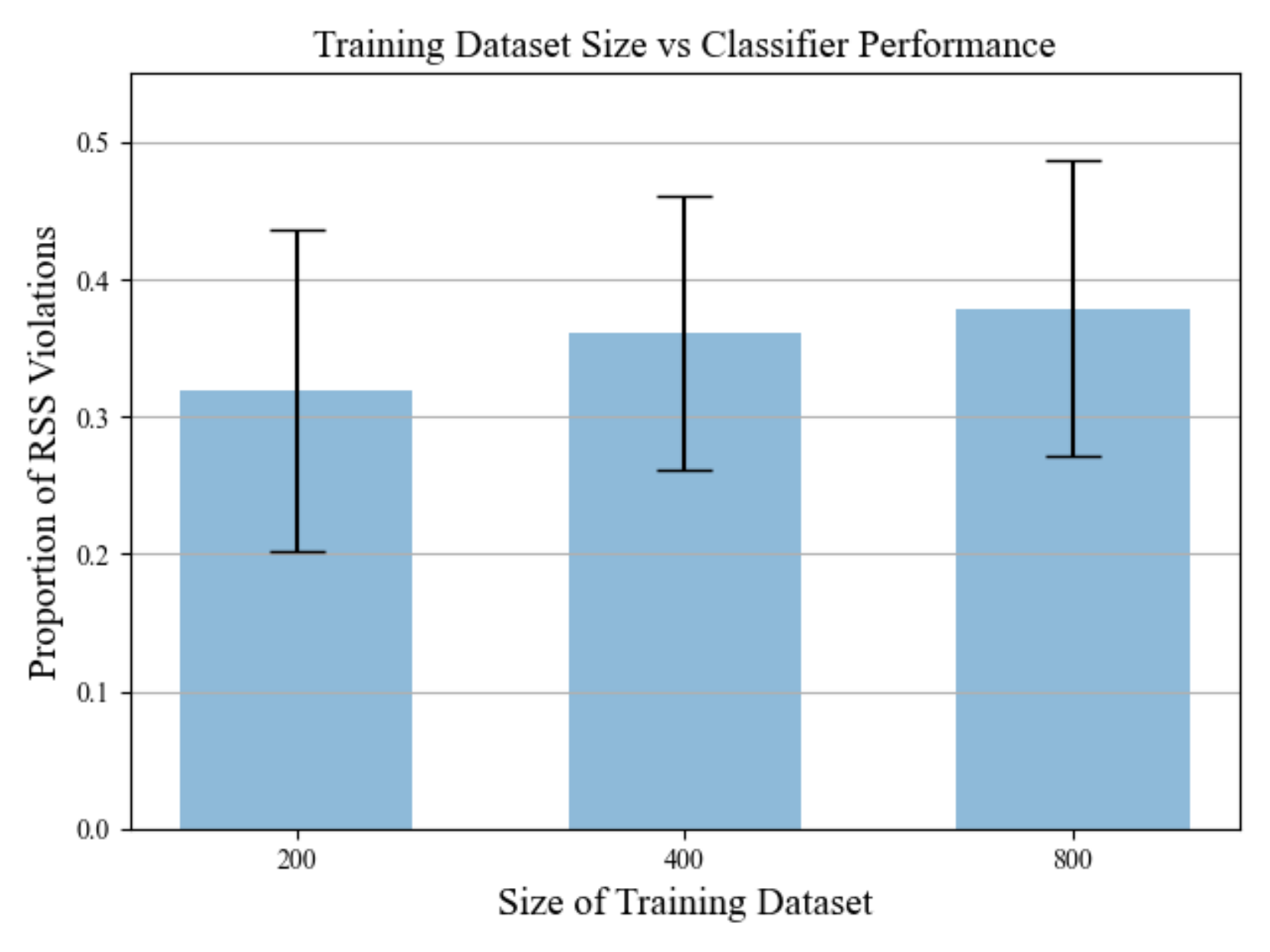}\\
    \caption{Performance of AST with HCS reward for classifiers trained with various dataset sizes.} 
    \label{fig:training_data_size}
    \vspace{-0.2cm}
\end{figure}

\subsection{Q-Value Critical State Reward}
AST with the Q-value critical state (QCS) reward was used to generate failure cases for the highway driving scenario. \cref{fig:RSS_hist_CS} shows the proportion of unsafe states according to RSS in failure trajectories found using the Q-value based critical state criteria ($R_{qcs}$). These are compared with failure trajectories obtained using the heuristic reward ($R_{heur}$). We see that such a definition results in a slight increase in the proportion of trajectories with RSS violations; however, the overall distribution remains comparable with two distinct peaks at similar locations. 

The method of classifying critical states using the variance of Q-values does not show a significant increase in the quality of failure trajectories found through AST. We observe that in the context of failure trajectory search, characterizing the reward based solely on the action-value pairs of the ego vehicle's controller is insufficient to identify a larger amount of actions (when compared to $R_{heur}$) that lead to dangerous behaviour. In many situations, the controller may have a small subset of significantly preferable actions, but these states fail to capture all scenarios where dangerous behaviour is exhibited.

\begin{figure}[b]
    \centering
    \hspace*{-1cm}
    \includegraphics[width=0.85\columnwidth]{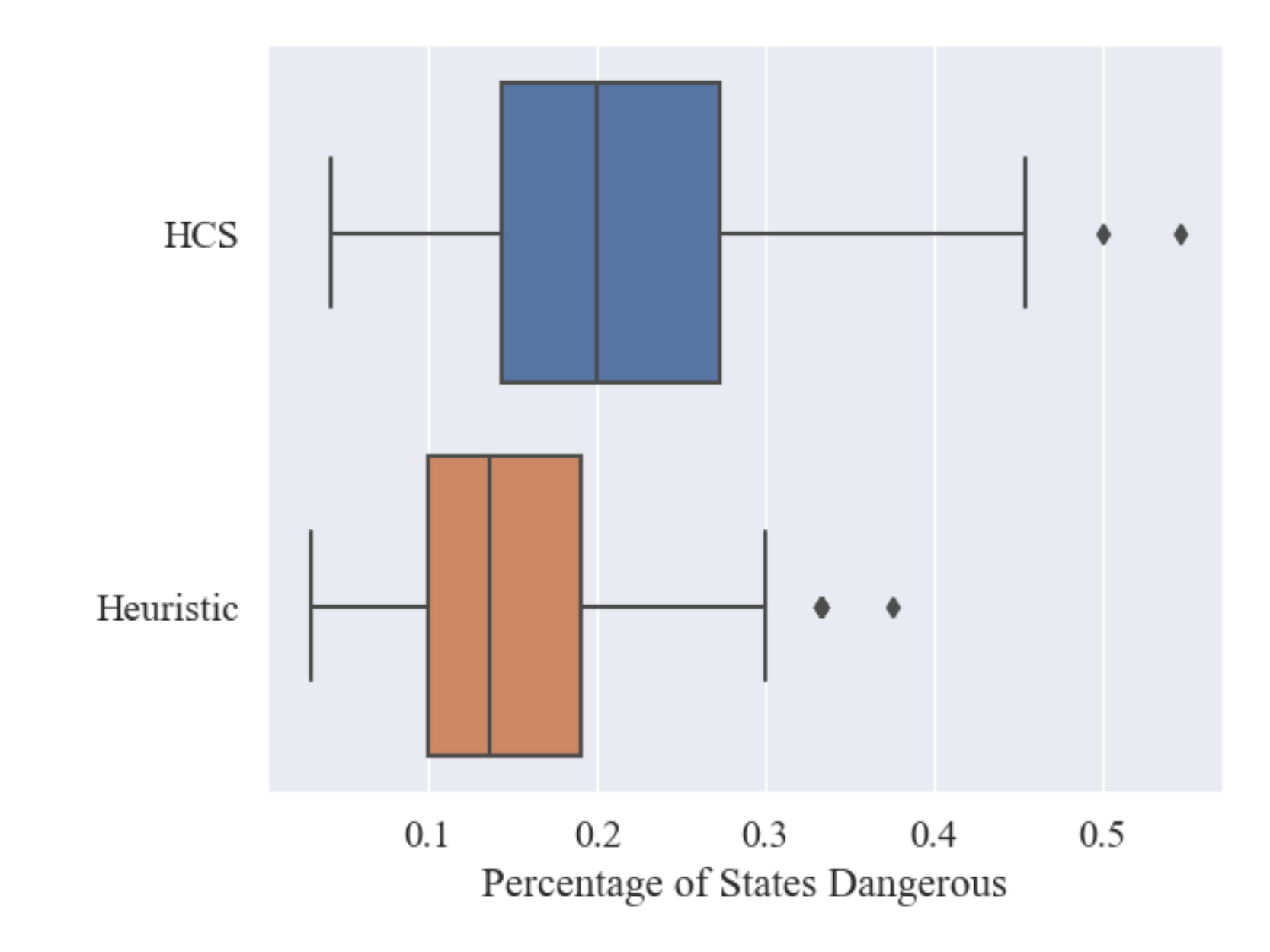}\\
    \caption{Box plot showing distribution of the percentage of states deemed critical (via human labelling) in failure trajectories found using HCS and heuristic rewards.}
    \label{fig:human_labelled_states_boxplot}
\end{figure}

\subsection{Human Critical State Reward}
We run AST failure search with the HCS reward on the same highway driving scenario and MCTS solver. \cref{fig:training_data_size} shows a comparison of AST performance with classifiers trained with increasing dataset sizes. By collecting training data for the classifier through both random simulation and AST with heuristic reward, we observe that the classifier converges in performance with a reasonably tractable amount of human labelled examples. 

\begin{figure}[h]
    \centering
    \begin{subfigure}{\columnwidth}
        \includegraphics[width=\columnwidth]{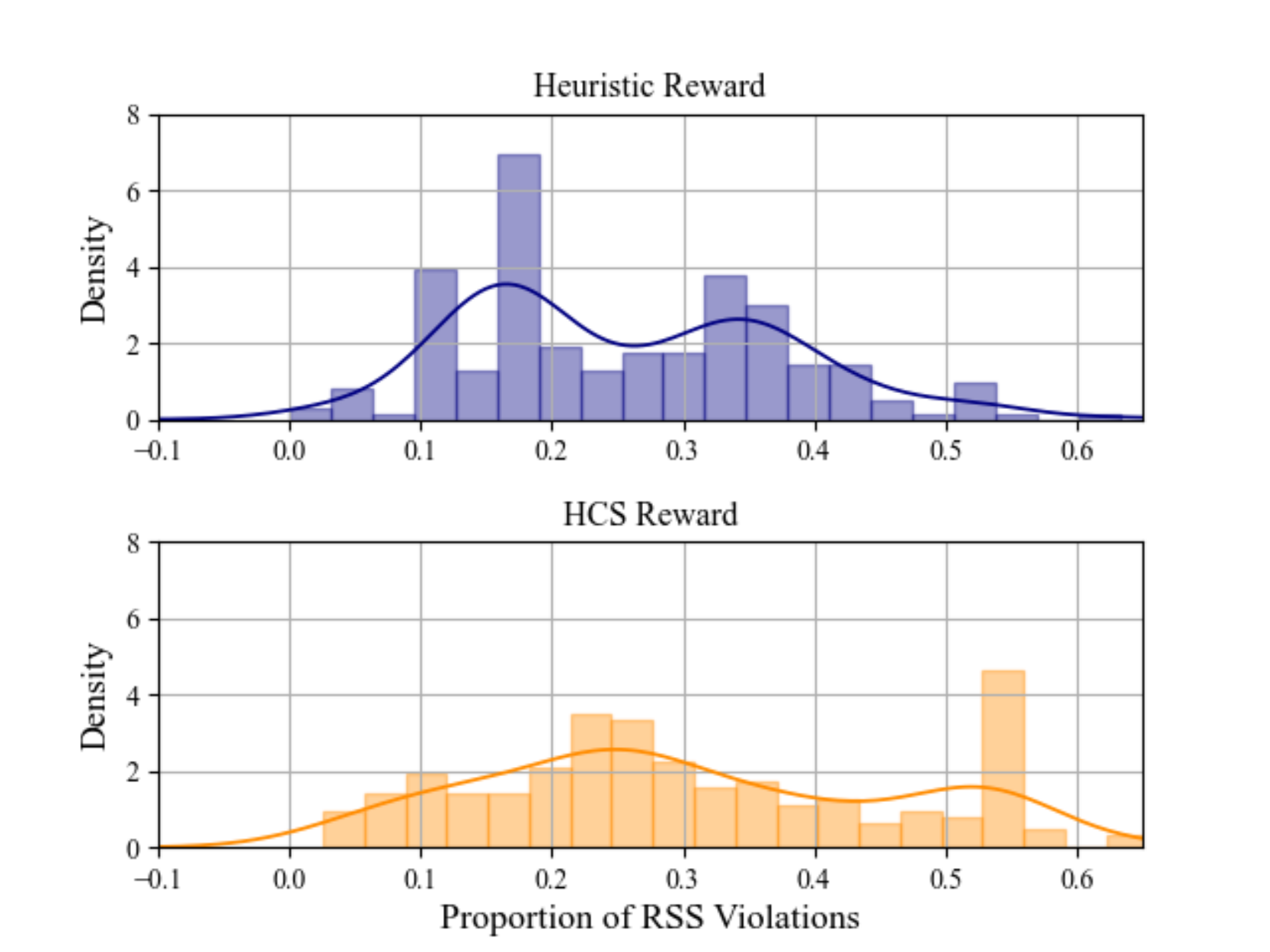}
        \caption{}
        \label{fig:RSS_hist1_a} 
    \end{subfigure}
    \vspace{0.25cm}
    \begin{subfigure}{\columnwidth}
        \includegraphics[width=\columnwidth]{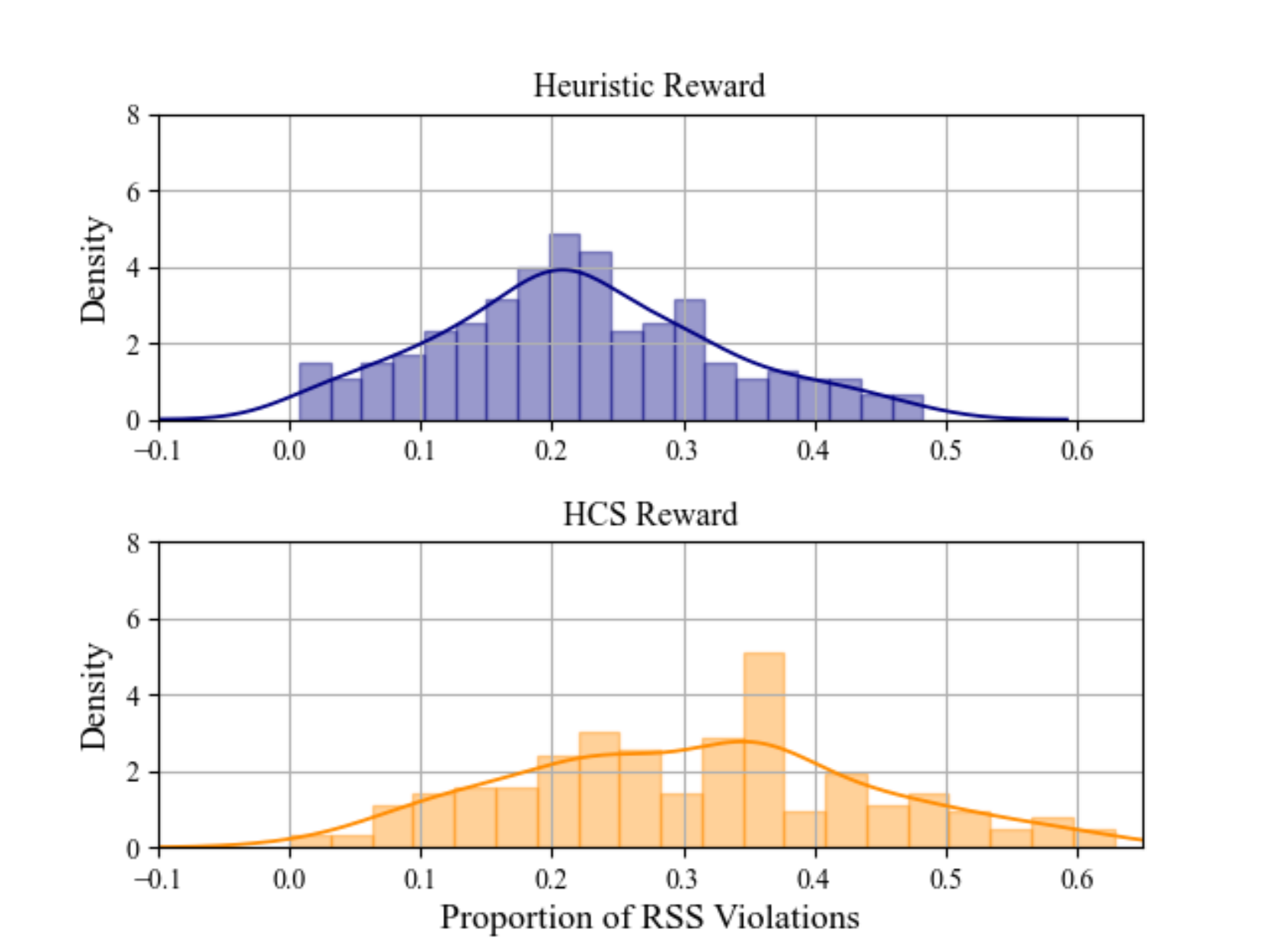}
        \caption{}
        \label{fig:RSS_hist1_b} 
    \end{subfigure}
    \caption{Normalized histograms of proportion of RSS violations for failure trajectories found with and without use of the HCS classifier. The graphs in (a) and (b) represent two different initial simulator configurations.}
    \label{fig:RSS_hist1} 
    \vspace{-0.3cm}
\end{figure}

As before, we can compare the results of failure trajectories found using the HCS reward with a baseline heuristic reward by looking at the proportion of states deemed unsafe by RSS. The initial conditions and simulator configuration is held constant between heuristic and HCS rewards. \cref{fig:RSS_hist1} shows the distribution of RSS violations for two distinct initial simulator states. We note that the use of a heuristic based on the longitudinal separation between the ego-vehicle and closest environment vehicle should be a good fit for the RSS criteria that is examined as it considers longitudinally unsafe scenarios. Despite this, we see a noticeable increase in the proportion of RSS violations using HCS reward. For example, the median proportion of improper response increases by $\sim 50\%$ from 21\% to 32\% in the second scenario (\cref{fig:RSS_hist1_b}).

In \cref{fig:RSS_hist1_a} the failure trajectories show an initial simulator configuration that leads to a multimodal distribution of RSS violations while \cref{fig:RSS_hist1_b} shows a configuration that leads to a unimodal distribution. In both cases, the use of the HCS reward results in a right shift in the modes of the distributions and AST identifying failures with a greater fraction of timesteps where RSS deems the scenario unsafe. The maximum fraction of states unsafe is also greater in both experiments using HCS reward. 

\begin{figure}[!t]
    \centering
    \hspace*{-0.2cm}
    \includegraphics[width=0.97\columnwidth]{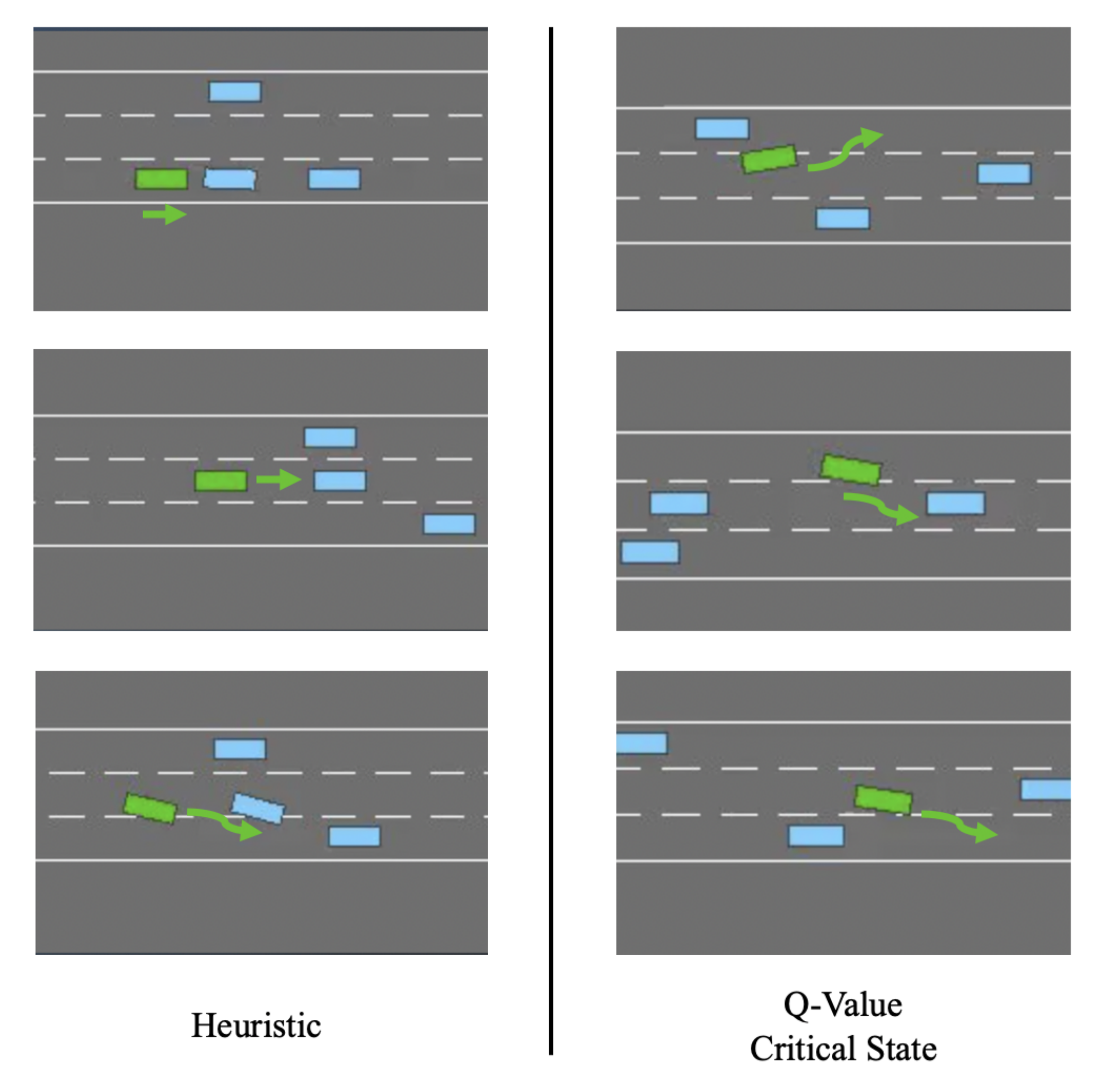}\\
    \caption{Critical scenarios of trajectory sampled from peak of RSS violations distribution for heuristic reward (left) and Q-value reward (right).}
    \label{fig:heuristic_scenarios} 
    \vspace{-0.15cm}
\end{figure}

The failure trajectories are human labelled to classify each state as either safe or unsafe. \cref{fig:human_labelled_states_boxplot} shows a boxplot with their distributions. In the case of HCS reward, we are directly optimizing to favour trajectories with an increased number of human deemed critical states and we see this reflected in the results. The use of HCS reward generates a more diverse set of failure trajectories with a greater spread over the percentage of states that exhibit dangerous behaviour.

\subsection{Qualitative Comparison}

\begin{figure}[t]
    \centering
    \hspace*{-0.15cm}
    \includegraphics[width=0.88\columnwidth]{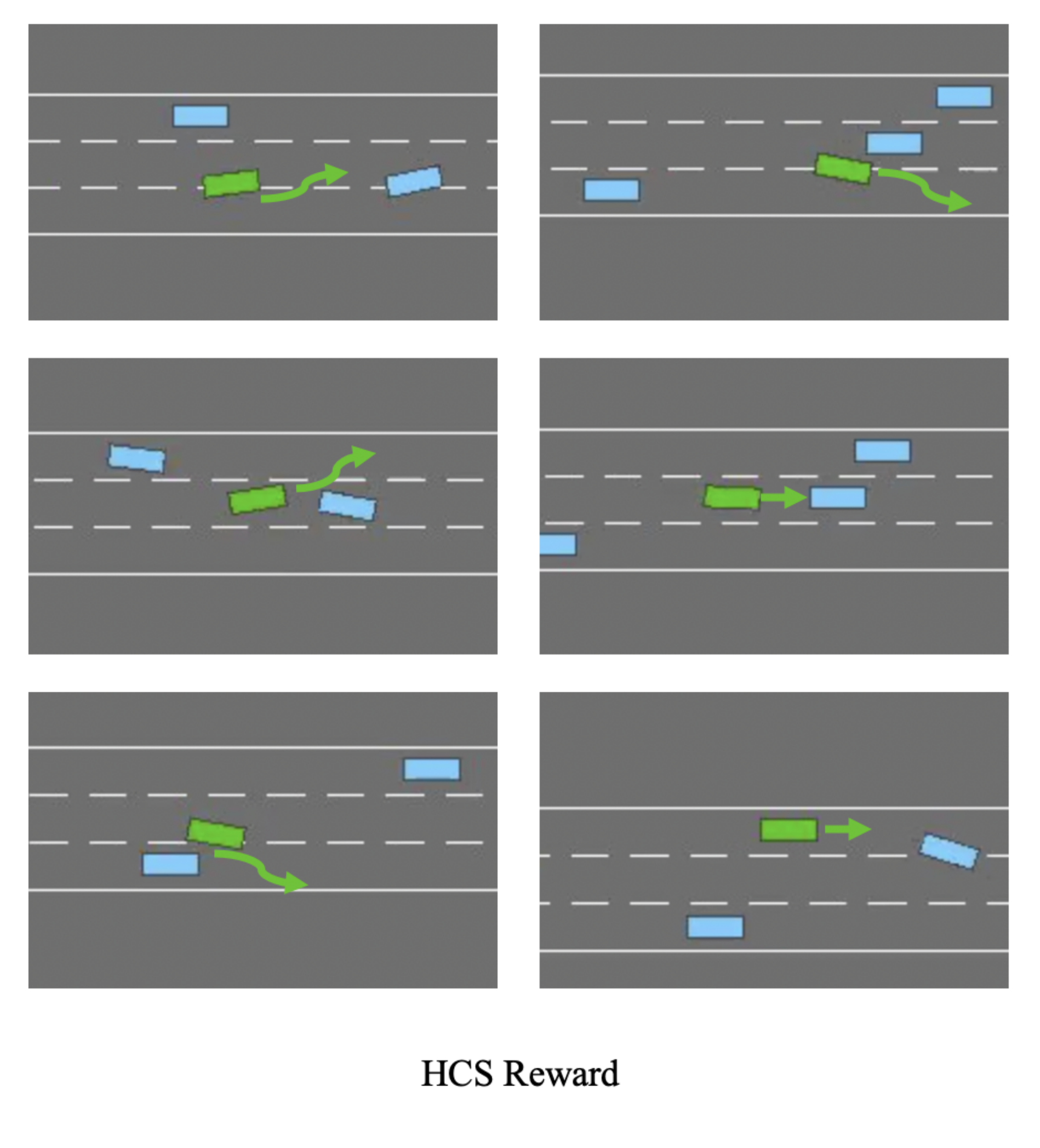}\\
    \caption{Critical scenarios of trajectory sampled from peak of RSS violations distribution for HCS reward.}
    \label{fig:hcs_scenarios} 
    \vspace{-0.15cm}
\end{figure}

We compare the results generated by the different reward formulations qualitatively. \cref{fig:heuristic_scenarios,,fig:hcs_scenarios} show all the instances of critical scenarios from failure trajectories found using heuristic, QCS, and HCS rewards respectively. Heuristic and HCS reward trajectories were sampled from the peaks of the RSS violations distributions show in \cref{fig:RSS_hist1_b}. The QCS reward trajectory was sampled from the peak of the distribution in \cref{fig:RSS_hist_CS}. The example from heuristic reward (\cref{fig:heuristic_scenarios} left) contains two types of dangerous behaviour: 1) The ego vehicle travels at a high rate of speed and fails to maintain a safe distance to the vehicle in front. 2) The ego vehicle makes a lane change where it loses a safe separation to the vehicle ahead of it. Critical scenarios found using QCS reward ((\cref{fig:heuristic_scenarios} right) show a similar result with the ego vehicle making an unsafe lane change in three separate instances (one loss of separation and two cut-offs where another vehicle is forced to brake abruptly). When compared with the two prior rewards, the failure trajectory sampled using HCS reward (\cref{fig:hcs_scenarios}) contains more critical scenarios and demonstrates both the unsafe following distance and lane change cut-off behaviour that is seen in heuristic and QCS rewards respectively. Qualitatively, HCS reward shows elevated levels of aggressive ego behaviour with smaller minimum separations and increased vehicle speeds.

\section{Conclusion}
\label{sec:conclusion}

\vspace{0.15cm}

This paper proposed the use of critical states in place of a probability proxy in the reward used to train a policy that generates failure scenarios of an autonomous agent using Adaptive Stress Testing. We presented the limitations of the traditional AST reward formulation for adaptive failure search in settings where the SUT and environment have a discrete action space representing high-level commands and the underlying distribution of environment actions is not readily available. We considered two characterizations of \textit{critical states}. The first defined critical states from the agent's perspective and characterized them using the variance of Q-values generated by an agent's policy. We showed that this formulation, while able to perform better than a purely heuristic reward, is not ideal for the task of failure search due to its limited ability to distinguish dangerous ego behaviour. This led us to characterize critical states as a classification problem that maps observations of the environment to criticality. We trained a Bayesian neural network to perform this task and demonstrated its effectiveness in generating failure trajectories with increased dangerous behaviour exhibited by the ego AV when used with AST. In the future, we hope to investigate more general feature encodings used to represent the state of the system when classifying critical states, such as the use of raw images of the environment to better incorporate spacial-temporal relations. 




\bibliographystyle{IEEEtran}
\bibliography{references}

\end{document}